# An Image Dataset of Common Skin Diseases of Bangladesh and Benchmarking Performance with Machine Learning Models

Sazzad Hossain[a], Saiful Islam[a], Muhammad Ibrahim[b,*], Md. Rasel Ahmed[a], Md Shuayb[c], Ahmedul Kabir[d]

[a]Department of Computer Science and Engineering, Faridpur Engineering College, Faridpur, Bangladesh

[b]Department of Computer Science and Engineering, University of Dhaka, Dhaka, Bangladesh

[c]Department of Oncology, Labaid Cancer and Super Speciality Center, Dhaka, Bangladesh

[d]Institute of Information Technology, University of Dhaka, Dhaka, Bangladesh

Emails: sazzadbytes@gmail.com (S. Hossain); saifulislamsayem19@gmail.com (S. Islam); ibrahim313@du.ac.bd (M. Ibrahim); rasel.ahmed@fec.ac.bd (R. Ahmed); shuayb@labaidcancer.com (M. Shuayb); kabir@iit.du.ac.bd (A. Kabir)

\* Corresponding author: Muhammad Ibrahim, Email: <u>ibrahim313@du.ac.bd</u>

## Abstract

Skin diseases are a major public health concern worldwide, and their detection is often challenging without access to dermatological expertise. In countries like Bangladesh, which is highly populated, the number of qualified skin specialists and diagnostic instruments is insufficient to meet the demand. Due to the lack of proper detection and treatment of skin diseases, that may lead to severe health consequences including death. Common properties of skin diseases are, changing the color, texture, and pattern of skin and in this era of artificial intelligence and machine learning, we are able to detect skin diseases by using image processing and computer vision techniques. In response to this challenge, we develop a publicly available dataset focused on common skin disease detection using machine learning techniques. We focus on five prevalent skin diseases in Bangladesh: Contact Dermatitis, Vitiligo, Eczema, Scabies, and Tinea Ringworm. The dataset consists of 1612 images (of which, 250 are distinct while others are augmented), collected directly from patients at the outpatient department of Faridpur Medical College, Faridpur, Bangladesh. The data comprises of 302, 381, 301, 316, and 312 images of Dermatitis, Eczema, Scabies, Tinea Ringworm, and Vitiligo, respectively. Although the data are collected regionally, the selected diseases are common across many countries especially in South Asia, making the dataset potentially valuable for global applications in machine learning-based dermatology. We also apply some machine learning and deep learning models on the dataset and report classification performance. We expect that this research would garner attention from machine learning and deep learning researchers and practitioners working in the field of automated disease diagnosis.

**Table 1: Dataset Specifications**

| Subject | Medical Image Classification |
|---|---|
| Specific subject area | Skin disease classification using deep learning |
| Type of data | Pictures saved digitally in the JPG format with dimensions 176 x 176 pixels and RGB colour space. |
| How the data were acquired | Skin disease data were collected directly from the outdoor department of Faridpur Medical College, Faridpur, Bangladesh with formal permission from the hospital administration. Images were taken using a mobile phone camera under natural lighting conditions, focusing on the affected skin areas. The dataset includes five categories: Contact Dermatitis, Tinea Ringworm, Eczema, Scabies, and Vitiligo. A total of around 250 raw images were collected before filtering and augmentation. |
| Data format | Filtered and pre-processed image dataset suitable for machine learning and deep learning model training. |
| Description of data collection | Images representing five common skin diseases were captured directly from patients visiting the aforementioned medical college. After obtaining administrative permission, each photo was taken with care to maintain image clarity and patient privacy. The images were resized, and low-quality samples were excluded. To improve dataset variety, simple augmentation methods like rotation and zoom were used. After processing and augmentation, the final dataset contains 302, 381, 301, 316, and 312 images of Dermatitis, Eczema, Scabies, Tinea Ringworm, and Vitiligo. |
| Data source location | The data were collected from the Dermatology Department, Faridpur Medical College, Faridpur, Bangladesh. Latitude and longitude: 23.5883374, 89.8327777 |
| Data Availability | https://data.mendeley.com/datasets/9ggd3shdr7/1 |

**Contributions**

A brief overview of this research's contributions is provided below:

- We develop a comprehensive, standard, and ready-to-use image dataset of skin diseases collected from patients of Bangladesh, addressing a critical gap in availability of dermatological data in the region. A total of 1612 images, of which 250 are distinct, are manually captured from the outpatient department of Faridpur Medical College, Bangladesh using standard imaging techniques.
- The dataset covers five common and clinically significant skin diseases prevalent in Bangladesh: Contact Dermatitis, Vitiligo, Eczema, Scabies, and Tinea Ringworm. These diseases are frequently observed and represent the majority of skin-related consultations in the country.

- We apply multiple data validation and preprocessing steps to ensure the quality and usability of the dataset. These include expert annotation by medical professionals, noise removal, ground-truth labeling, image resizing, and data augmentation (including zooming and rotation) to enhance model generalization.
- The dataset is released for public use and is readily available for download, enabling researchers to directly integrate the images into machine learning pipelines without the need for additional preprocessing or validation.
- Although the dataset is region-specific, the selected skin conditions are widespread globally, especially in south Asia. Therefore, this dataset is well-suited as pre-trained models for transfer learning applications and can contribute to the development of generalized AI models for skin disease detection in low-resource or underserved regions worldwide.
- We apply several machine learning and deep learning models on the dataset and achieve benchmarking performance.

**Objective**

A highly impactful area of current research is the application of machine learning, especially deep learning models in healthcare, particularly for automating the detection of skin diseases through medical images. Reliable and representative datasets are essential to train and evaluate these models effectively. However, our survey of existing literature and public datasets reveals a significant gap — there is a lack of standard, high-quality datasets that focus specifically on common skin diseases prevalent in Bangladesh and similar regions. These datasets are often limited in size, class balance, image quality, and expert labeling [15, 27, 28].

In response to this need, in this study, our objective is to develop and release a standard, ready-to-use dataset of skin disease images collected from real clinical settings. This dataset focuses on five commonly encountered diseases in Bangladesh, namely Contact Dermatitis, Vitiligo, Eczema, Scabies, and Tinea Ringworm. By creating and publicly sharing this dataset, we aim to foster research in machine learning-based dermatological diagnostics and promote the broader use of AI for healthcare solutions in resource-constrained settings.

Our initiative is grounded in the belief that the full potential of machine learning and deep learning has not yet been realized in critical domains such as healthcare. By bridging the data gap, especially in underrepresented regions, we hope this effort contributes, even modestly, toward democratizing AI and making its benefits accessible to the masses.

**Data Description**

To accurately predict the labels of dataset instances using a Convolutional Neural Network (CNN), each class must have discrete visual features that enable CNNs to extract distinctive features. These inter-class differences allow the model to effectively learn and distinguish between categories during training and prediction. In this section, we examine the unique visual characteristics of various skin diseases represented in our dataset images, highlighting the features that support reliable classification. The five diseases—Contact Dermatitis, Vitiligo, Eczema, Scabies, and Tinea Ringworm—exhibit noticeable visual differences in terms of skin color, texture, shape, and lesion pattern. These differences are essential for the model to learn meaningful representations and perform accurate classification. Figures 1, 2, 3, 4, 5 and 6 show two sample images of each of the five skin disease classes included in our dataset and two healthy skin images.

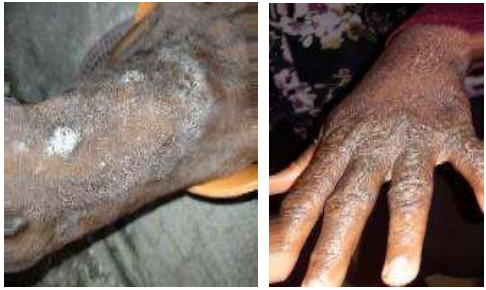

Figure 1: Contact Dermatitis

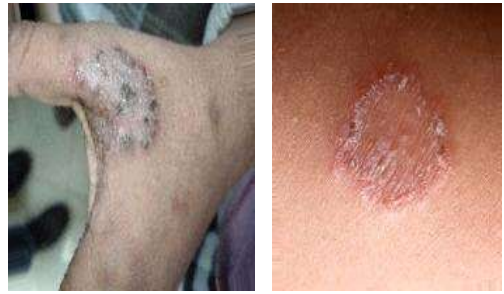

Figure 2: Tinea Ringworm

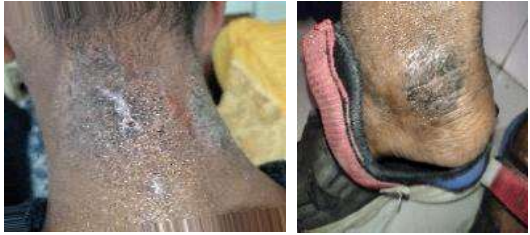

Figure 3: Eczema

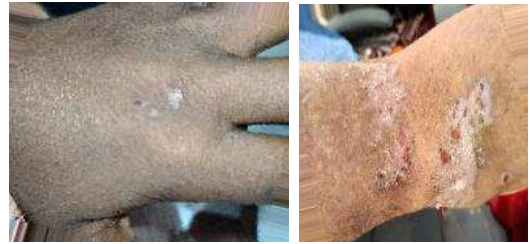

Figure 4: Scabies

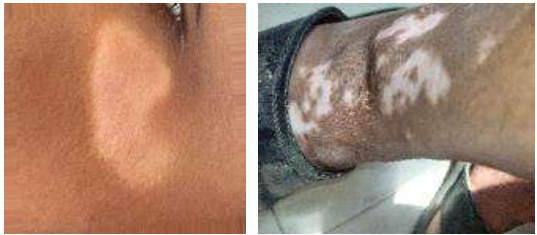

Figure 5: Vitiligo

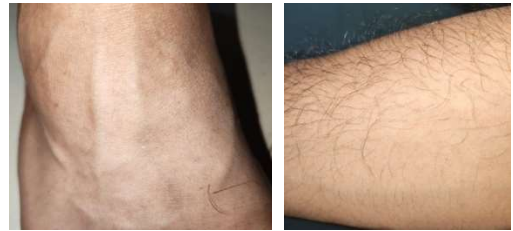

Figure 6: Healthy skin

**Contact Dermatitis** appears as a red, itchy rash often caused by direct contact with an irritant or allergen. The affected area may become swollen, dry, cracked, or blistered, and may ooze or crust over time; see **Figure 1**. The rash usually follows a pattern based on the shape or location of exposure.

**Tinea (Ringworm)** is a fungal skin infection marked by a red, circular rash with clearer skin in the middle. The border of the lesion may be raised, scaly, or blistered; see **Figure 2**. Despite its name, ringworm has no relation to worms and can affect the scalp, body, feet, or groin.

**Eczema** (Atopic Dermatitis) is a chronic condition where red, inflamed, itchy patches develop on the skin. These patches may thicken, crack, or scale over time due to persistent scratching; see **Figure 3**. It most commonly affects the face, neck, and folds of the elbows or knees in children, but can occur anywhere on the body.

**Scabies** is a contagious skin condition caused by a mite infestation, leading to intense itching and a pimple-like rash. Small burrow tracks can often be seen in the folds of skin—especially between fingers, around the waist, or on wrists; see **Figure 4**. The itching tends to worsen at night.

**Vitiligo** is characterized by smooth, depigmented white patches on the skin due to the loss of melanocytes. These patches can appear on any part of the body, including the face, hands, and joints, and often grow larger with time; see **Figure 5**. The affected skin maintains its texture but lacks pigment, creating a sharp contrast with surrounding normal skin.

In contrast to the abovementioned diseases, **healthy skin** is uniform in color and texture, free from visible rashes, lesions, or dryness. It appears smooth, supple, and well-hydrated without any flaking, redness, or irritation; see **Figure 6**. Healthy skin does not exhibit abnormal pigmentation or texture changes.

The above discussion shows that each skin condition exhibits unique visual patterns, making this domain well-suited for deep learning-based classification and diagnosis.

**Experimental Design, Materials and Methods**

Dermatologists and other health care providers have historically used visual examination to diagnose skin conditions. However, recent quick technological advancements have made it possible to use digital tools to help diagnose a variety of skin conditions more quickly and affordably. Machine learning is unique among these cutting-edge technologies in that it can accurately predict disease labels when enough pertinent historical data are available. Deep learning algorithms are currently being successfully used in a variety of sectors such as agriculture, social welfare, including healthcare, by utilizing low-priced computing resources [1, 2, 6, 13]. A high-quality dataset is essential for successful machine learning applications. The process of building or training the model involves these algorithms extracting hidden patterns from the dataset, which they then use to predict future labels. Factors like the size of the dataset, the balance of class distributions, between-class distinctiveness, within-class consistency, and the lack of noise in the data and labels can all be used to assess the quality of dataset. Datasets with these desirable characteristics are regarded as standard. Our goal in this study is to develop a standardized, usable, publicly accessible image dataset of five prevalent skin conditions in Bangladesh. Although the skin is one of the most visible organs in the body, there are currently no region-specific, clinically annotated image datasets available in public repositories of Bangladesh for conditions like contact dermatitis, vitiligo, eczema, scabies, and tinea ringworm.

**Methodology**

In deep learning, the importance of data cannot be overstated. In is widely known in machine learning practitioners that the quality and quantity of data often have a greater impact on model performance than the specific algorithms used. Therefore, it is essential for researchers to follow a set of best practices throughout the dataset preparation process. These practices typically include selecting representative real-world samples, cleaning the data, augmenting it, and applying accurate labels.

In this section, we first outline the steps we followed in preparing our dataset. Next, we examine the visual features of skin images associated with various diseases. Finally, we discuss the major challenges encountered during the development of the dataset.

**Steps for Dataset Collection and Preparation**

The dataset preparation process involved the following key phases:

- Gained a thorough understanding of common skin diseases in Bangladesh to guide data collection.

- Formed partnerships with dermatology experts and selected appropriate clinical sites for data collection.
- Selecting patient cohorts and designing data collection protocols so that the data collection and usage procedures meet standard ethical practices and guidelines.
- Captured images of diseased skin areas from patients. A total of five skin diseases that are commonly found in patients of Bangladesh were considered.
- Image validation and processing: This stage included:
    - manual labeling of images by human experts,
    - resizing images to a standardized format commonly used in machine learning,
    - cleaning images to remove background noise, and
    - applying data augmentation techniques such as zooming and rotation.

Below we elaborate each of these four phases:

**Examining Common Skin Conditions.** Common skin diseases like Contact Dermatitis, Vitiligo, Eczema, Scabies, and Tinea (Ringworm) are visually identifiable and significantly affect health and well-being. Contact Dermatitis appears as red, defined patches [5], Vitiligo as white depigmented spots [8], [18], Eczema as itchy, scaly plaques [14], Scabies as burrows and papules [9][16], and Tinea as ring-shaped lesions with central clearing [12]. These diseases are widespread, with Eczema affecting up to 20% and Tinea causing 20–25% of fungal infections [10], highlighting the need for early detection. Machine learning-based image processing techniques have shown strong potential in diagnosing these conditions. A CNN-based system for Ghana achieved over 84% accuracy for multiple diseases [1], and reviews show deep learning often outperforms traditional methods [2]. Techniques like SVM, KNN, and hybrid models have been shown to reach up to 98% accuracy in various studies [3], [4]. Hospital data from Bangladesh confirm the high burden of these diseases, stressing the need for improved diagnostic tools [5]. Advanced models using CNNs, DenseNet, and InceptionV3 have achieved up to 99.2% accuracy, supporting fast and reliable skin disease detection [14], [20], [21].

**Selecting the Clinical Site and Designing Data Collection Protocols.** For data collection, we partnered with the outpatient department of Faridpur Medical College in Faridpur district of Bangladesh. This site was strategically chosen due to its high patient volume and the wide variety of skin conditions treated, thereby ensuring a rich and diverse dataset. The selection also aimed to reduce sampling bias by capturing cases from a broad demographic. Over a six-month period, the clinical site provided consistent exposure to both common and complex skin diseases that confirms its suitability for this study.

**Physically Capturing the Images.** Skin photographs were taken immediately after dermatological consultation under natural outdoor lighting. A Redmi Note 10 (48-megapixel) captured images in both portrait and landscape orientations at resolutions up to 4000×3000 pixels. There was negligible delay between consultation and imaging, ensuring minimal lesion alteration. Temperature and humidity reflected typical local conditions, with neither overcast nor foggy weather. This process yielded 1,612 raw images across the six categories.

**Validating the Images of the Dataset.** To standardize inputs, each image was resized to 176×176 pixels and saved as JPG. Minor artifacts were manually removed, and severely blurred or occluded images were discarded, resulting in 250 high-quality images. Data augmentation is a well-known and effective practice in machine learning community. We then applied zoom and rotation augmentations to reach a total of 1,612 images. Images showing multiple disease traits were omitted to preserve clear interclass distinctions. All labels were corroborated by dermatology experts, ensuring reliable ground truth.

Figure 9 represents the flowchart for the data preparation processes. Table 1 shows all of the dataset's information at a glance.

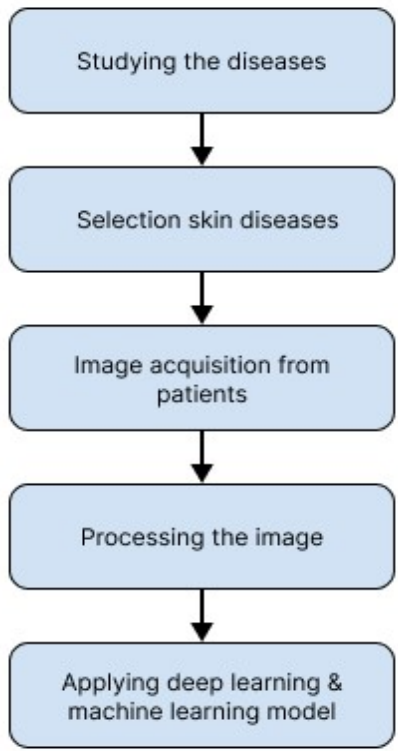

Figure 9: Flowchart showing the data preparation steps.

Table 1: Proposed skin disease dataset information at a glance.

| Type of data: | 176x176 Clinical skin images (external body parts only). |
|---|---|
| Data format | JPG. |
| Number of images | 1612 pictures. About 250 of these are of unique skin photos, while others are generated by rotating and zooming as needed. |
| Diseases considered | Five commonly found diseases: Contact Dermatitis, Vitiligo, Eczema, Scabies, Tinea Ringworm |
| Number of classes | Five (five disease categories) |
| Distribution of instances | - Contact Dermatitis: 302<br>- Eczema: 381<br>- Scabies: 301<br>- Tinea Ringworm: 316<br>- Vitiligo: 312 |

| How data are acquired | Photographed by mobile phone immediately after outpatient consultation. |
|---|---|
| Data source locations | Outdoor area, Faridpur Medical College Dermatology department, Faridpur, Bangladesh |
| Where applicable | Suitable for healthy vs. diseased binary classification and multi-class disease prediction. |

**Difficulties with Skin Image Dataset Preparation**

Building a machine learning dataset from the ground up is a difficult and time-consuming process that calls for a great deal of human commitment and effort. Nevertheless, the time and money spent eventually pay off because a well-curated dataset can help countless researchers and machine learning experts when it is made publicly available.

The main difficulties we ran into when preparing our dataset. Firstly, Diversity of skin diseases: even among our five target conditions—Contact Dermatitis, Vitiligo, Eczema, Scabies, and Tinea Ringworm—there is wide variability in lesion appearance, location, and severity. Secondly, Clinical site logistics: obtaining sufficient images for less common conditions (e.g. Vitiligo) required extended time in the outpatient department of Faridpur Medical College and careful scheduling with dermatology staff. Thirdly, Technical difficulties during image acquisition: capturing consistent, high-resolution photographs with a mobile phone under varying outdoor lighting and ensuring minimal patient discomfort was challenging. Fourthly, many choices of data validation techniques: deciding which preprocessing and augmentation methods (resizing, normalization, noise reduction, color balancing, zoom/rotation) would best enhance model performance required substantial expertise in both data science and dermatology.

**Applied Machine Learning Models and Experimental Results**

To establish benchmarking performance on the newly developed dataset, we apply several traditional machine learning models and a deep learning (CNN) model on our dataset. Below we briefly describe them.

**Traditional Machine Learning Models**

**a. Naive Bayes**
We use Gaussian Naive Bayes that is trained and evaluated using a grid search approach for hyperparameter tuning. The grid search is conducted with 5-fold cross-validation. Finally, we find the best Gaussian Naive Bayes parameters as {*var_smoothing*: 1e-09} with an accuracy score of 62.81%.

**b. The Support Vector Machine (SVM)**
For SVM classifier, we set the parameter grid with a specified combination of hyperparameters indicating a radial basis function (RBF), a linear kernel with gamma and a regularization parameter set to 5. The grid search is conducted with 5-fold cross-validation to evaluate each parameter combination. The best SVM parameters are found as: {*C*: 1, gamma: 0.1, *kernel*: linear} with an accuracy score of 95.04%.

**c. K-Nearest Neighbors (KNN)**

Our KNN classifier also adopts a grid search approach for hyperparameter tuning. In KNN the only hyperparameter being tuned is *n_neighbors*, which represents the number of neighbors to consider. Initially, we set it to 5 and other hyperparameters, such as the distance metric, the weight function. The grid search is conducted with 5-fold cross-validation. The best parameter is found to be 3 with an accuracy score of 91.74%.

**d. The Gradient Boosting**

For gradient boosting classifier, we set the parameter grid with various hyperparameters such as n_estimators, learning_rate, max_depth, max_features, loss, and subsample, each with a list of values to explore during the grid search. The grid search is conducted with 5-fold cross-validation to evaluate each parameter combination. We find the best parameters: {*learning_rate*: 0.01, *loss*: log_loss, *max_depth*: 5, *max_features*: log2, *n_estimators*: 150} with an accuracy score of 95.04%.

**e. Random Forest**

The Random Forest classifier is trained and evaluated using a grid search approach for hyperparameter tuning. The grid search is conducted with 5-fold cross-validation to evaluate each parameter combination. The best parameters are found to be: {*max_depth*: 5, *max_features*: log2, *min_samples_leaf*: 1, *min_samples_split*: 2, *n_estimators*: 200} with an accuracy score of 83.88%.

Fig. 10 summarizes the performances of all these five models both on training and test sets.

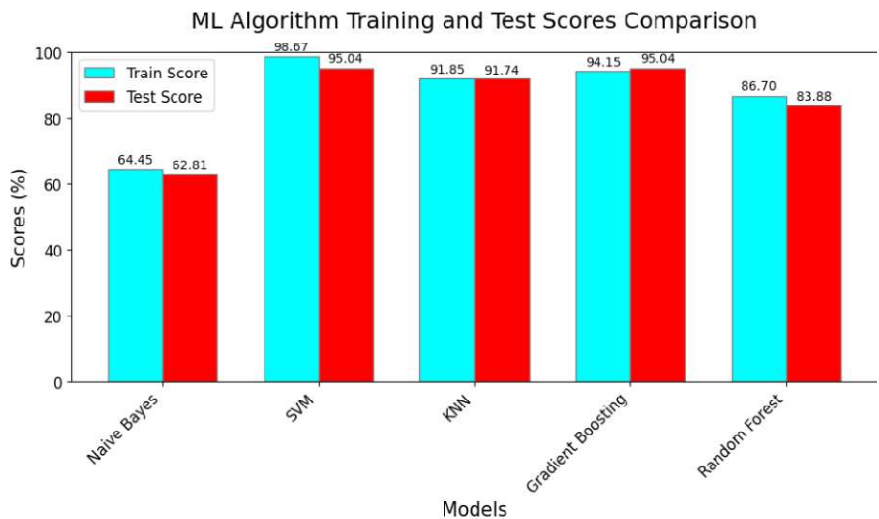

Figure. 10: Traditional machine learning algorithms: training and test accuracy scores.

**Deep Learning Model**

For CNN model building, we use the sequential model implying the layers are added one by one in sequence. It is made up of several convolution and pooling layers, each of which serves a particular function in feature extraction and hierarchical learning. The input shape is (176, 176, 3), corresponding to the height, width, and number of channels (RGB) of the input images. We use two fully connected layers, one with 1024 neurons and another with 512 neurons. The final output layer has 5 neurons, which is equal to the number of classes in the classification task. It uses the softmax activation function to output class probabilities. Finally, we train the model on the training data for a specified number of epochs while validating on the validation data. The training set contains 70% of the data and the remaining 30% is allocated for testing and validation equally. We also set early stopping and learning rate reduction callbacks to monitor the validation loss and accuracy during training and perform necessary actions.

A bar chart is given Fig. 11 for these five diseases, which gives us a clear picture of the accuracy of each disease for our CNN model.

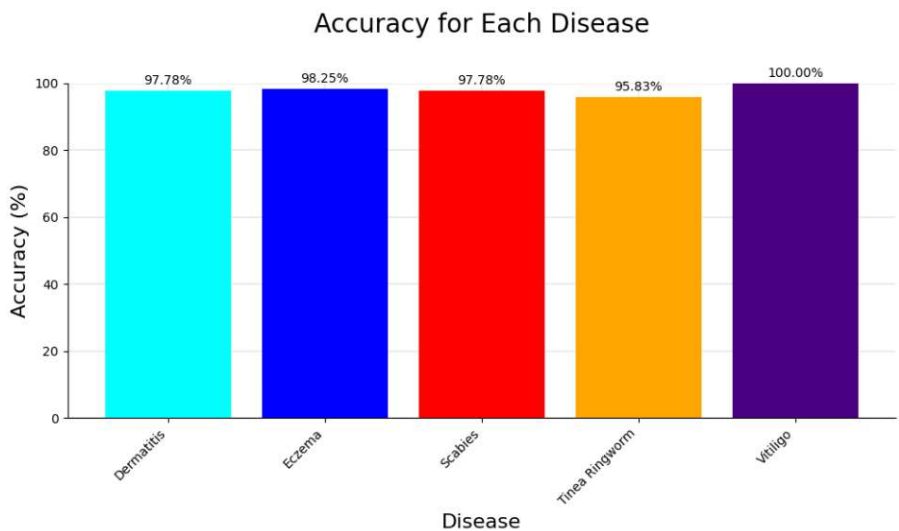

Figure 11. Accuracy for each disease class using CNN model.

**Confusion Matrices**

In this section, we analyze the confusion matrices of all the six algorithms.

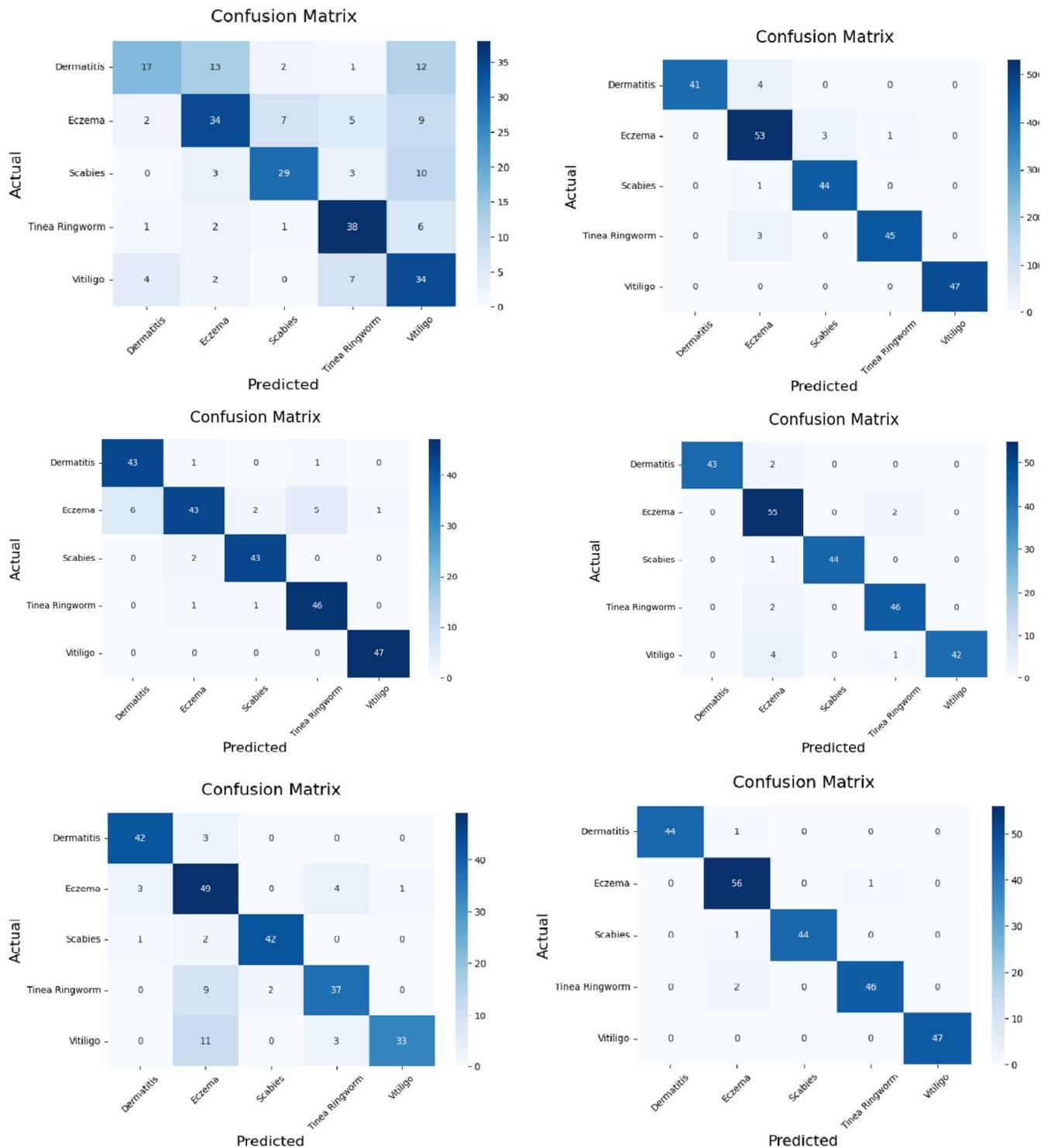

Figure 12. Confusion matrices (from left to right, top to bottom) of naïve bayes, SVM, kNN, Gradient boosting, random forest, and CNN algorithms.

For naïve bayes (Fig. 12, top-left), we see that in the Dermatitis class, only 17 images are predicted correctly as Dermatitis and most of the images are predicted wrongly in other classes. In Eczema class, 34 images are predicted correctly as Eczema.

As well, Scabies has 29 correct predictions, Tinea Ringworm has 38 correct predictions and Vitiligo has 34 correct predictions.

For SVM (Fig. 12, top-right), we see that in Dermatitis class, 41 images are predicted correctly as dermatitis and four images are predicted wrongly as eczema. In Eczema class, 53 images are predicted correctly as Eczema and 3 images are predicted wrongly as Scabies, and one as Tinea Ringworm. As well, Scabies has 44 correct predictions with 1 wrong prediction and Tinea Ringworm has 45 correct predictions with 3 wrong predictions. But in VVitiligo, all 47 images are predicted correctly without any wrong predictions.

For KNN (Fig. 12, middle-left), we see that in Dermatitis class, 43 images are predicted correctly as Dermatitis and one image is predicted wrongly as eczema and one as Tinea Ringworm. In Eczema class, 43 images are predicted correctly, while others are predicted wrongly. As well, Scabies has 43 correct predictions with two wrong prediction and Tinea Ringworm has 46 correct predictions with two wrong predictions. But in Vitiligo, all 47 images are predictedcorrectly without any wrong predictions.

For gradient boosting (Fig. 12, middle-right), we see that in the Dermatitis class, 43 images are predicted correctly as dermatitis and two images are predicted wrongly as eczema. In Eczema class 55 images are predicted correctly as Eczema and two images are predicted wrongly. As well, Scabies has 44 correct predictions with one wrong prediction and Tinea Ringworm has 46 correct predictions with two wrong predictions. In Vitiligo, 42 images are predicted correctly and five are predicted wrong.

For random forest (Fig. 12, bottom-left), we see that in the Dermatitis class, 42 images are predicted correctly as dermatitis and three images are predicted wrongly as eczema. In Eczema class, 49 images are predicted correctly as Eczema and others are predicted wrongly in other classes. As well,Scabies has 42 correct predictions with three wrong predictions and Tinea Ringworm has 37 correct predictions. At last, in Vitiligo, 33 images are predicted correctly, while others are predicted wrongly in other classes.

For CNN (Fig. 12, bottom-right), we see that in the Dermatitis class, 44 images are predicted correctly as dermatitis and only one image is predicted wrongly as eczema. In Eczema class, 56 images are predicted correctly as Eczema and only one image is predicted wrongly as Tinea ringworm. As well as in Scabies has 44 correct predictions with 1 wrong prediction and Tinea Ringworm has 46 correct predictions with 2 wrong predictions. But in Vitiligo all 47 images are predicted correctly without any wrong predictions.

**Conclusion**

This research has developed and released a standard, ready-to-use skin disease image dataset by capturing images from outdoor patients of a medical college of Bangladesh. We have also applied several machine learning and deep learning algorithms on this dataset and achieved satisfactory accuracy. The CNN model achieved the highest accuracy of 97.93%. Our experimental results show that the findings of this research will pave the way to automate the skin disease detection of under-developed countries like Bangladesh, thereby minimizing the cost of providing healthcare to mass people.

**Acknowledgment**

This research is supported by Dhaka University Research Grant, 2024-25, provided by the University of Dhaka, Dhaka-1000, Bangladesh. We express gratitude to this entity.